\pdfoutput=1
\documentclass[10pt, logo, copyright]{nvidiatechreport}
\usepackage{caption}
\usepackage{mdframed}
\usepackage[utf8]{inputenc} 
\usepackage[T1]{fontenc}    
\usepackage{hyperref}       
\usepackage{url}            

\usepackage{booktabs}       
\usepackage{amsfonts}       
\usepackage{nicefrac}       
\usepackage{microtype}      
\usepackage[dvipsnames]{xcolor}         
\usepackage{multirow}
\usepackage{multicol}
\usepackage{graphicx}
\usepackage{subcaption}
\usepackage[numbers]{natbib}
\usepackage{tabto}
\usepackage{xspace}
\usepackage{amsmath}
\usepackage{adjustbox}
\usepackage{enumitem}
\usepackage{makecell}
\usepackage{array}
\newcolumntype{N}{>{\centering\arraybackslash}p{1.2cm}} 
\usepackage{wrapfig}
\usepackage{dblfloatfix}
\usepackage{algorithm}
\usepackage{algpseudocode}
\usepackage{comment}
\usepackage{cuted}
\usepackage[most]{tcolorbox}
\usepackage{xcolor}
\usepackage{graphicx}
\usepackage{hyperref}
\usepackage{xcolor}

\newcolumntype{g}{>{\color{gray}}c} 

\newcommand{\VS}{\mathcal{V}_{S}}
\newcommand{\VT}{\mathcal{V}_{T}}
\newcommand{\MatchSet}{\mathcal{M}}
\newcommand{\CandidateSet}{\mathcal{A}}
\newcommand{\ChunkSet}{\mathcal{C}}
\newcommand{\Div}{\mathcal{D}}
\newcommand{\Dec}{\operatorname{Dec}}

\newcommand{\method}{\textsc{CSD}\xspace}

\title{CSD: Common-Support Distillation for Cross-Tokenizer Knowledge Transfer}
\author{
Mingyu Yang,
Sharath Turuvekere Sreenivas,
Adithyakrishna Venkatesh Hanasoge,
Nima Tajbakhsh,
Ashwath Aithal,
Pavlo Molchanov
}

\begin{abstract}
\textbf{Abstract:}
Cross-tokenizer logits distillation enables knowledge transfer across model families, allowing students to learn from stronger or more specialized teachers. However, differences in vocabulary and segmentation make teacher and student next-token distributions difficult to align. Existing logits-based methods are easy to integrate into standard distillation pipelines, but often rely on restrictive heuristics or tokenizer-pair-specific choices. They may restrict supervision to shared vocabulary subsets, require manually selecting different objectives for different tokenizer pairs, or match only the observed chunk and its complement rather than
individual alternative continuations. We introduce Common-Support Distillation (\method), a unified logits-based framework that constructs a shared support between the teacher and student at each aligned input chunk and matches the induced distributions with a single objective. For one-to-one alignments, \method uses explicit common-token pairs as the shared support. For other alignments, it constructs a virtual support whose probabilities are computed using the autoregressive chain rule. A residual event captures probability mass outside the explicit support, ensuring that both induced distributions remain normalized. Across four teacher–student settings comprising Llama-3.2-1B and Gemma3-1B-PT students with Qwen and Phi teachers, \method consistently achieves the highest average performance among the evaluated cross-tokenizer methods, outperforming the strongest alternative by up to $1.8$ points and surpassing both continued pretraining and same-tokenizer distillation. For Llama-3.1-8B, \method improves average accuracy over the strongest evaluated baseline by up to $4.2$ points for the base student and $1.6$ points for the instruction-tuned student. 
\end{abstract}

\begin{document}
\maketitle


\vspace{-0.1in}
\section{Introduction}
\label{sec:introduction}


Knowledge distillation (KD) transfers knowledge from a teacher to a smaller student, typically by matching their predictive distributions~\citep{hinton2015distilling}. In language models, standard token-level KD assumes a shared vocabulary and aligned prediction positions. When teacher and student use different tokenizers, the same text is segmented differently which requires token level alignment, and their distributions assign probability
mass to different token outcomes. Direct distribution matching therefore requires both aligning prediction contexts and establishing corresponding outcomes across vocabularies. These requirements complicate token-level KD and restrict the choice of teachers.

We tackle the issue of logits distillation between models of different families and different tokenizers. Existing approaches differ in how they handle vocabulary and segmentation mismatches. GOLD~\citep{patino2025_unlocking_on_policy_distillation_for_any_model_family} directly matches teacher and student probabilities for tokens with identical strings in both vocabularies. For tokens without such a match, it sorts each model's remaining probabilities and compares them by rank, ignoring token identity. X-Token identifies two limitations of this partition: loss of task-critical token identity and suppressive gradients on unmatched tokens~\citep{sreenivas2026x}. Its P-KL and H-KL modes use projection and relaxed hybrid matching, respectively. However, selecting between them relies on a manual coverage audit for each tokenizer pair. Approximate Likelihood Matching (ALM) provides a uniform formulation across tokenizer pairs by matching the probability assigned to each observed aligned chunk and aggregating all remaining probability mass into a single residual event~\citep{minixhofer2025universal}. While broadly applicable, this aggregation discards fine-grained dark knowledge about how the teacher distributes probability among alternative continuations.

We introduce Common-Support Distillation (\method), which makes teacher and student predictions comparable by organizing them over a shared support of text-equivalent continuation events. We first align token chunks that decode to the same observed text span. For one-to-one chunks, the support consists of common-token pairs from the two vocabularies. For one-to-many or many-to-one chunks, each event pairs a single token from one model with a token sequence from the other. For example, given the observed alignment $(\texttt{120}\leftrightarrow\texttt{[1][2][0]})$, fixing the prefix \texttt{[1][2]} and varying the final token yields
$\texttt{120}\leftrightarrow\texttt{[1][2][0]},\ldots, \texttt{129}\leftrightarrow\texttt{[1][2][9]}$. We fix the within-chunk prefix to \texttt{[1][2]} because the
forward pass on the observed sequence provides the conditional
distributions along this prefix. A sequence event is assigned the product of its constituent conditional token probabilities. For accepted many-to-many chunks, the observed alignment forms the sole explicit event. Finally, a residual event captures probability mass outside the selected support, yielding normalized distributions that \method matches with a single objective across all chunk types and tokenizer pairs.

Our contributions are summarized as follows, and the implementation of \method is publicly available\footnote{\url{https://github.com/NVIDIA-NeMo/RL/pull/4680}}.
\begin{itemize}[leftmargin=*,nosep]
\item \textbf{A uniform cross-tokenizer objective.}
\method matches teacher and student distributions over a shared
support of paired token sequences plus a residual event. The same
distribution-matching objective applies across chunk shapes and tokenizer pairs without
pair-specific loss selection.
\item \textbf{Consistent gains across model families.}
Across four settings Llama-3.2-1B and Gemma3-1B-PT students and
Qwen3-4B and Phi-4-mini teachers, \method achieves the highest average accuracy,
outperforming X-Token by $0.66$--$5.38$ points and ALM by
$1.33$--$3.30$ points, while also surpassing continued pretraining and
same-tokenizer KD.

\item \textbf{Efficient top-$k$ distillation.}
In the Qwen--Llama setting, using only the top $16$ or $128$
teacher tokens reduces the five-task average by just $1.74\%$ or
$0.74\%$ relative to full logits, substantially lowering
training-step time with minimal accuracy loss.

\item \textbf{Gains at larger scales and training stages.}
For the Llama-3.1-8B base student, \method outperforms continued pretraining by
$2.17$--$4.43$ points and ALM by $2.61$--$4.23$ points across
two teachers. For Llama-3.1-8B-Instruct, it improves over
continued training and ALM by $3.90$ and $1.60$ points.

\end{itemize}

\vspace{-0.1in}
\section{Related Work}
\label{sec:related-work}
Existing cross-tokenizer distillation methods differ in how they make teacher and student predictions comparable. We group these approaches into representation-based methods and logits-based methods.
\vspace{-0.2in}
\paragraph{Representation-based methods.}
These methods access hidden states to align different output spaces. Dual-Space Knowledge Distillation (DSKD) learns projections between teacher and student representations and uses cross-model attention to align sequences before comparing predictions~\citep{zhang2024dual}.  Byte-Level Distillation (BLD) attaches a byte-level prediction head to student hidden states and trains it using teacher probabilities mapped to byte-level outcomes~\citep{singh2026cross}.
\vspace{-0.2in}
\paragraph{Logits-based methods.}
These methods derive supervision from existing output logits, using tokenizer information for alignment.  Several approaches transfer probabilities through token correspondences: MinED uses edit-distance alignment~\citep{wan2024knowledge}; Contextual Dynamic Mapping (CDM) refines sequence and vocabulary mappings using contextual predictions~\citep{chen2025enhancing}; and SeDi combines graph-based token-span alignment, vocabulary re-encoding, and entropy alignment~\citep{wang2026bridging}.  Other approaches change the quantities being compared.  Universal Logit Distillation (ULD) matches sorted probability vectors, discarding token identity~\citep{boizard2024towards}, and MultiLevelOT extends this
optimal-transport formulation to both the token and sequence
levels~\cite{cui2025multilevelot}. GOLD utilizes a hybrid loss formulation where KL is used for common tokens in both vocabularies and ULD loss for tokens without such a match~\citep{patino2025_unlocking_on_policy_distillation_for_any_model_family}. X-Token addresses limitations of GOLD through two loss modes: projection (P-KL) and relaxed hybrid (H-KL) matching. However, it requires a manual coverage audit to decide the loss mode for each tokenizer pair~\citep{sreenivas2026x}.
Approximate Likelihood Matching (ALM) instead uses a common chunk-level formulation across tokenizer pairs, matching distributions over the observed aligned chunk and its complement~\citep{minixhofer2025universal}. However, it does not individually match the probabilities of alternative tokens. 
\vspace{-0.2in}
\paragraph{Cross-tokenizer on-policy distillation.}
A recent line of work extends these ideas to on-policy distillation (OPD), where the
teacher scores trajectories generated by the student. Beyond GOLD, DPCA aligns
student and teacher sequences into minimal synchronized chunks for on-policy
training~\cite{niu2026dpca}. SimCT compares the two models on short
continuations that both tokenizers can express, supervising only the positions
and candidates where such continuations exist~\cite{sun2026simct}. Byte-Prefix
Marginalization (BPM) re-expresses the teacher's next-token distribution over
the full student vocabulary by routing each teacher token's probability to the
longest student token whose bytes prefix it, with unmatched mass assigned to a
residual category~\cite{wang2026bpm}.

\method shares with ALM and BPM the use of a residual event to keep the matched distributions normalized. It differs in how the explicit
support is formed. ALM matches only the observed chunk, and BPM maps teacher
mass into the student vocabulary through byte-prefix routing. \method instead
constructs a shared support on both sides: common-token pairs with identical
decoded text for one-to-one chunks, and chain-rule events over alternative
final tokens for one-to-many and many-to-one chunks. This symmetric
construction requires only text equivalence and makes no assumption about how
either tokenizer segments a candidate continuation. Although we study CSD in
the off-policy setting, its per-chunk construction depends only on the observed
text and both models' predictions, so it applies unchanged to on-policy
trajectories.

\vspace{-0.1in}
\section{Common-Support Distillation}
\label{sec:method}
\label{sec:problem}

\begin{figure*}[t]
  \centering
  \includegraphics[width=\linewidth]{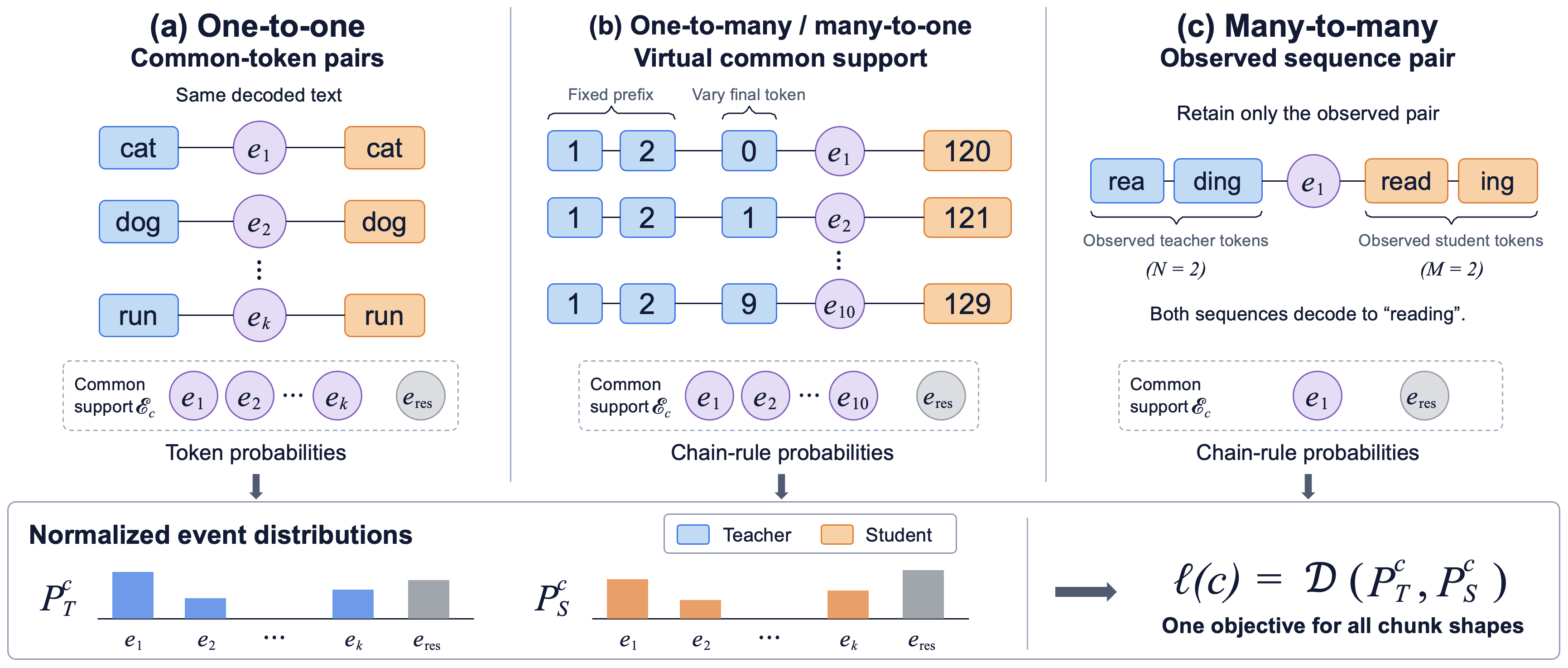}
  \vspace{-0.15in}
  \caption{Common-support construction in \method, illustrated with example
tokenizations. (a) One-to-one chunks use common-token pairs. (b) In the illustrated one-to-many chunk, the teacher prefix \texttt{[1][2]} is fixed while the final token varies, matching single student tokens \texttt{120}--\texttt{129}. Swapping the model roles gives the many-to-one case.
(c) Many-to-many chunks retain only the observed sequence pair.
A residual event preserves all remaining mass, and the same objective
matches the resulting normalized distributions. Tokenizations and
probability bars are illustrative.}
  \label{fig:common-support-overview}
  \vspace{-0.15in}
\end{figure*}

\method matches teacher and student predictions of corresponding textual outcomes, conditioned on the same preceding text. It comprises three stages. First, sequence alignment identifies paired token ranges covering
the same substring using character offsets and decoded-text checks. Second, common-support construction maps each chunk's predictions onto shared events defined by common-token pairs or token sequences representing the same text (Figure~\ref{fig:common-support-overview}). Event probabilities are computed directly from token predictions or through the autoregressive chain rule. A residual event retains all remaining mass, ensuring normalization. Finally, distribution matching applies a uniform objective across chunk shapes and teacher--student tokenizer pairs. Algorithm~\ref{alg:csd-training} in Appendix~\ref{app:csd-pseudocode} summarizes the computation of the \method training objective for one batch.
\vspace{-0.1in}
\subsection{Sequence alignment}
\label{sec:aligned-spans}
\label{sec:predictions}

\paragraph{Token-level predictions.}
Let a text sequence $x$ be tokenized by the student and teacher as
  $\mathbf{s}=(s_1,\ldots,s_{n_S}), \;
  \mathbf{t}=(t_1,\ldots,t_{n_T}),$
with vocabularies $\VS$ and $\VT$, respectively. Their next-token
distributions are denoted by $p_S(\cdot\mid\mathbf{s}_{<i})$ over $\VS$ and
$p_T(\cdot\mid\mathbf{t}_{<j})$ over $\VT$. In this case, even when $i=j$, the histories may represent different text. Moreover, their distributions assign probabilities to different vocabulary tokens. Thus, direct token-level matching is not suitable here.
\vspace{-0.2in}
\paragraph{Aligned chunks.}
We first define $(i,j)$ as a \emph{common boundary} if the histories
$\mathbf{s}_{<i}$ and $\mathbf{t}_{<j}$ cover the same prefix of the original text. Then, given two
consecutive common boundaries $(i,j)$ and $(i+M,j+N)$, with $M,N\geq1$, we can
define an \emph{aligned chunk} $c=(\mathbf{s}^{c},\mathbf{t}^{c})$ covering
the intervening substring, where
$
    \mathbf{s}^{c}
    =(s_i,\ldots,s_{i+M-1})=(s_1^c,\ldots,s_M^c), \;\;
    \mathbf{t}^{c}
    =(t_j,\ldots,t_{j+N-1})=(t_1^c,\ldots,t_N^c).
$
The shape $M{:}N$ records the student and teacher token counts in the aligned chunk. We refer to chunks with $M=N=1$ as \emph{one-to-one} and to chunks with $M>1$ or $N>1$ as \emph{non-one-to-one}.
In cross-tokenizer distillation, the aim is to match the probabilities of corresponding student and teacher realizations within each aligned chunk.
\vspace{-0.2in}
\paragraph{Finding aligned chunks with character offsets.}
For student tokens $s_r$ and teacher tokens $t_r$, their tokenizers provide character spans $[a_r^S,b_r^S)$ and
$[a_r^T,b_r^T)$ in the original text. Thus, we use a two-pointer scan that starts from tokens
with the same start offset and extends whichever range ends earlier until
the end offsets first coincide. That is, for an aligned chunk $c=(\mathbf{s}^{c},\mathbf{t}^{c})$, the two token ranges share the same character span
$[a_c,b_c)$, with 
$a_c:=a_i^S=a_j^T$, $b_c:=b_{i+M-1}^S=b_{j+N-1}^T$, and $a_c<b_c$.
The retained set $\ChunkSet$ excludes unmatched ranges and candidates that
fail the decoded-chunk checks, which allow Unicode normalization and
specified whitespace equivalences. We provide the full procedure for offset-based alignment in Appendix~\ref{app:alignment}.
\vspace{-0.1in}
\subsection{Common-support construction}
\label{sec:common-support}

Alignment identifies the observed token ranges and their histories. We next define the events whose probabilities are matched within each aligned chunk. This event set forms the common support $\mathcal{E}_c$, constructed as shown in Figure~\ref{fig:common-support-overview}. For one-to-one chunks,
each explicit event pairs student and teacher realizations of the same decoded text. Each retained token-pair set contains no repeated student or
teacher token IDs. For non-one-to-one chunks, we build virtual events whose probabilities can be computed for both models. These events, together with a residual event, define the common support for the distributions $P_S^c$ and $P_T^c$ compared in distillation.
\vspace{-0.1in}
\subsubsection{One-to-one chunks}
\label{sec:matching-boundaries}
\paragraph{Common-token pairs.}
For one-to-one chunks (Figure~\ref{fig:common-support-overview}(a)), the common support can be built by checking the token pairs in the student vocabulary $\VS$ and teacher vocabulary $\VT$ that decode to the same text fragment. For $u\in\VS$ and $v\in\VT$, we define the common-token pairs as:
$
  \MatchSet=\{(u_k,v_k)\}_{k=1}^{K}
  \subseteq \VS\times\VT,
$
where $K$ denotes the total number of retained token pairs. This relation is shared across one-to-one chunks for a tokenizer pair and includes alternatives beyond the observed tokens.
\vspace{-0.2in}
\paragraph{Normalized event distributions.}
Given the token pairs, each event $e_k^c$ pairs student token $u_k$
with teacher token $v_k$, while $e_{\mathrm{res}}^c$ collects
unpaired tokens. The common support is
$
  \mathcal{E}_c=\{e_1^c,\ldots,e_K^c,e_{\mathrm{res}}^c\}.
$
Let $h_S^c=\mathbf{s}_{<i}$ and
$h_T^c=\mathbf{t}_{<j}$ denote the student and teacher token
histories preceding chunk $c$. These histories represent the
same preceding text.
In this event order, their distributions can be represented as
$
P_S^c = \Bigl(
p_S(u_1\mid h_S^c),\ldots,p_S(u_K\mid h_S^c), 1-\sum_{k=1}^{K}p_S(u_k\mid h_S^c)
\Bigr), \;\;
P_T^c = \Bigl(
p_T(v_1\mid h_T^c),\ldots,p_T(v_K\mid h_T^c), 1-\sum_{k=1}^{K}p_T(v_k\mid h_T^c)
\Bigr).
$
The residual incorporates unmatched tokens through their aggregate probability mass, avoiding potentially inaccurate token-level correspondences as in GOLD~\citep{patino2025_unlocking_on_policy_distillation_for_any_model_family}. X-Token~\citep{sreenivas2026x} identifies suppression in GOLD's original common-token loss: a common-token-only loss computed from
full-vocabulary softmax probabilities pushes unmatched student logits
downward. For that loss, including the residual makes the gradient
depend on the teacher--student residual-mass mismatch, allowing
either upward or downward pressure. Thus, even a task-critical token is not always penalized
for lacking a common-token counterpart. ALM uses the observed chunk as its sole explicit event~\citep{minixhofer2025universal}, whereas \method expands this to include alternative common tokens (Appendix~\ref{app:method-comparison}).
\vspace{-0.1in}
\subsubsection{Non-one-to-one chunks}
\label{sec:mismatched-boundaries}

\paragraph{Virtual common support.}
When segmentation differs, corresponding text requires multiple tokens on
at least one side. We therefore construct a virtual common support from
paired token sequences. For a chunk of shape $M{:}N$ with
$(M,N)\neq(1,1)$, we fix the observed prefixes $\mathbf{s}_{<M}^c$ and
$\mathbf{t}_{<N}^c$ and allow only the final tokens to vary. The prefixes
contain $M-1$ and $N-1$ tokens, respectively, and are empty on a
single-token side ($M=1$ or $N=1$).

Let $\Dec_S$ and $\Dec_T$ denote the decoding functions and
$\mathbin{\Vert}$ token-sequence concatenation. A final-token pair
$(u,v)\in\VS\times\VT$ defines a \emph{paired completion} if
$
  \Dec_S(\mathbf{s}_{<M}^c\mathbin{\Vert}u)
  =
  \Dec_T(\mathbf{t}_{<N}^c\mathbin{\Vert}v).
$
The paired sequences must decode to the same text, which need not equal the
observed chunk text; their token counts remain $M$ and $N$. Let the retained valid final-token pairs form the candidate set $\CandidateSet_c=\{(u_k^c,v_k^c)\}_{k=1}^{K_c}$,
with no repeated student or teacher token IDs.
Each shared event $e_k^c$ pairs
$\mathbf{s}_{<M}^c\mathbin{\Vert}u_k^c$ and
$\mathbf{t}_{<N}^c\mathbin{\Vert}v_k^c$.

We define shared events this way, with fixed prefixes of $M-1$ and $N-1$ tokens, so that their probabilities can be computed from the observed-prefix predictions using the chain rule:
\begin{equation}
  \begin{aligned}
    q_S^c(k)
    &=
      \left[
        \prod_{r=1}^{M-1}
        p_S(s_r^c\mid h_S^c\mathbin{\Vert}\mathbf{s}_{<r}^c)
      \right]
      p_S(u_k^c\mid h_S^c\mathbin{\Vert}\mathbf{s}_{<M}^c),\\
    q_T^c(k)
    &=
      \left[
        \prod_{r=1}^{N-1}
        p_T(t_r^c\mid h_T^c\mathbin{\Vert}\mathbf{t}_{<r}^c)
      \right]
      p_T(v_k^c\mid h_T^c\mathbin{\Vert}\mathbf{t}_{<N}^c).
  \end{aligned}
  \label{eq:chain-probabilities}
\end{equation}
An empty product equals one. Each $q_S^c(k)$ and $q_T^c(k)$ is the probability of a specific token-sequence realization, computed by the chain rule without length normalization.

For one-to-many or many-to-one chunks ($M=1$ or $N=1$, Figure~\ref{fig:common-support-overview}(b)), one side represents each candidate text with a single token. We can therefore precompute bidirectional mappings between student and teacher tokens, enabling $O(1)$ lookup per mapping entry when retrieving valid final-token pairs. In contrast, for many-to-many chunks ($M,N>1$, Figure~\ref{fig:common-support-overview}(c)), valid pairs depend on the observed prefixes on both sides, making exhaustive precomputation impractical. Constructing a larger support therefore requires additional chunk-specific matching. To avoid this cost, we retain only the observed final-token pair $(s_M^c,t_N^c)$. This is a computational choice rather than a
requirement of the objective. We evaluate richer many-to-many
support in Section~\ref{sec:mismatch-support}.
\vspace{-0.2in}
\paragraph{Normalized event distributions.}
Each pair defines an event $e_k^c$ with probabilities $q_S^c(k)$ and
$q_T^c(k)$. In the order $(e_1^c,\ldots,e_{K_c}^c,e_{\mathrm{res}}^c)$,
$
    P_m^c=
    \left(
      q_m^c(1),\ldots,q_m^c(K_c),
      1-\sum_{k=1}^{K_c}q_m^c(k)
    \right), \;m\in\{S,T\}.
$
Within each model, candidates share a fixed prefix
and have distinct final tokens. They therefore define mutually
exclusive token-sequence events whose total probability is at
most one, ensuring a nonnegative residual. For accepted many-to-many chunks,
we have $K_c=1$, resulting in a binary distribution similar to ALM~\citep{minixhofer2025universal}. For one-to-one, one-to-many, and many-to-one with $K_c>1$, the expanded support provides finer-grained supervision by preserving the relative probabilities of multiple alternative events. Appendix~\ref{app:support-size-distribution} reports the empirical support-size distributions.
\vspace{-0.1in}
\subsubsection{Top-$k$ support selection}
\label{sec:topk-support}

\method is compatible with top-$k$ selection on either or both models, or directly over paired events. We can treat top-$k$ selection as restricting the candidate vocabulary used for common-support construction at each prediction position. For student position $i$ and teacher position $j$, define
$
    \mathcal{V}_{S,k}^{i}
    =\operatorname{TopK}_{u\in\VS}
      p_S(u\mid\mathbf{s}_{<i}), \;\;
    \mathcal{V}_{T,k}^{j}
    =\operatorname{TopK}_{v\in\VT}
      p_T(v\mid\mathbf{t}_{<j}),
$
where $\operatorname{TopK}$ returns the set of $k$ highest-probability tokens. 

The reduced vocabularies restrict which token pairs enter the
common support. For one-to-one chunks, only common-token pairs
whose tokens remain in the candidate vocabularies contribute.
For non-one-to-one chunks, selection restricts the
final-token pairs at student position $i+M-1$ and teacher position
$j+N-1$. The observed within-chunk prefixes remain fixed, and
their conditional probabilities must also be available for
chain-rule scoring. Thus, top-$k$ selection reduces the explicit
common support.

Retained probabilities come from the original full-vocabulary distributions
without renormalization. Excluded mass is absorbed by the existing residual
event on each side, so both event distributions remain normalized and the
matching objective retains the same. Our experiments evaluate
teacher-side top-$k$ in Section~\ref{sec:topk-efficiency}.
\vspace{-0.1in}
\subsection{Distribution matching}
\label{sec:divergence}

Both single-token and multi-token constructions yield normalized event
distributions $P_T^c$ and $P_S^c$, allowing us to use one matching rule.
Although support sizes may differ across chunks, the teacher and student
distributions refer to the same events within each chunk.
A retained chunk is \emph{accepted} for KD if its support contains at least
one explicit event ($K_c \geq 1$). Chunks that do not meet this criterion contribute no KD term. 
\vspace{-0.2in}
\paragraph{Per-chunk loss.}
For an accepted chunk $c$, we use
$\ell(c)=\Div(P_T^c,P_S^c),$
where $\Div$ can be any suitable divergence, including forward KL (FKL),
reverse KL (RKL), or Jensen--Shannon divergence (JSD). The loss compares the probabilities of all
explicit events together with the residual mass, retaining aggregate
supervision for outcomes outside the explicit support.
The same selected $\Div$ applies across chunk shapes and tokenizer pairs,
without pair-specific loss selection.
\vspace{-0.2in}
\paragraph{Batch loss.}
Let $\ChunkSet_{1:1}$ and $\ChunkSet_{\ne}$ be the accepted one-to-one and
non-one-to-one chunks in a batch. Each accepted chunk contributes
one loss term. For a nonempty accepted set, we aggregate these terms as
$
  \mathcal{L}_{\mathrm{KD}}^{\beta,\Div}
  =
    (\sum_{c\in\ChunkSet_{1:1}}\ell(c)
    +\beta\!\sum_{c\in\ChunkSet_{\ne}}\ell(c) )
  /
    (|\ChunkSet_{1:1}|+|\ChunkSet_{\ne}|)
  .
$
The configurable weight $\beta\geq0$ controls the relative contribution of
non-one-to-one chunks. It weights their losses without changing the underlying
support. The denominator counts accepted chunks
independently of $\beta$. If none is accepted, we set
$\mathcal{L}_{\mathrm{KD}}^{\beta,\Div}=0$.

\vspace{-0.2in}
\paragraph{Dynamic joint training.}
Following X-Token~\citep{sreenivas2026x}, we combine the student's
next-token cross-entropy $\mathcal{L}_{\mathrm{CE}}$ with dynamically
rescaled KD. Cross-entropy supervises the observed student tokens, while
KD transfers the teacher's probabilities over the shared events.
For $\mathcal{L}_{\mathrm{KD}}^{\beta,\Div}>0$,
$
  \mathcal{L}_{\mathrm{train}}
  =
  \operatorname{sg}\!\left(
    \mathcal{L}_{\mathrm{CE}} /
         \mathcal{L}_{\mathrm{KD}}^{\beta,\Div}
  \right)
  \mathcal{L}_{\mathrm{KD}}^{\beta,\Div}
  +\mathcal{L}_{\mathrm{CE}},
$
where $\operatorname{sg}$ stops gradients through the entire ratio.
The scaling factor is computed from the current batch losses and treated
as a constant during backpropagation. This balances scalar loss values,
not necessarily gradient magnitudes. Student gradients flow through all
retained chain factors and explicit and residual probabilities; the
teacher remains fixed. If KD is zero, training uses
$\mathcal{L}_{\mathrm{CE}}$ alone.
\vspace{-0.1in}
\section{Experiments}
\label{sec:experiments}
\vspace{-0.1in}
\subsection{Experimental Setup}
\label{sec:experimental-setup}

\begin{table*}[t]
  \caption{Main results for the Llama-3.2-1B student. Bold and underlined values denote the best and second-best scores within each cross-tokenizer teacher group.
    $^\dagger$ X-Token configurations are selected using tokenizer-pair coverage analysis.}
  \label{tab:llama1b-main}
  \centering
  \scriptsize
  \setlength{\tabcolsep}{2pt}
  \renewcommand{\arraystretch}{0.95}
  \resizebox{\linewidth}{!}{%
  \begin{tabular}{ll|*{5}{N}|N}
    \toprule
    \textbf{Setting} & \textbf{Method}
      & \textbf{MMLU} & \textbf{WG}
      & \textbf{GSM8K} & \textbf{MATH} & \textbf{MBPP} & \textbf{Avg.-5} \\
    \midrule
    \multirow{2}{*}{No distillation}
      & Llama-3.2-1B (base) & 32.05 & 61.48  & 5.69 & 5.48 & 26.40 & 26.22 \\
      & Continued pretraining & 42.14 & 62.27 & 14.71 & 7.82 & 27.00 & 30.79 \\
    \midrule
    Same tokenizer
      & Llama-3.2-3B  & 43.94 & 62.75 & 12.28 & 7.76 & 28.20 & 30.99 \\
    \midrule
    \multirow{5}{*}{\makecell[l]{Cross-tokenizer\\Qwen3-4B}}
      & ULD & 39.19 & 61.96 & 12.74 & 6.40 & 25.60 & 29.18 \\
      & ALM & 42.81 & \textbf{62.98} & \underline{16.07} & \textbf{8.74} & 29.00 & 31.92 \\
      & GOLD & \textbf{45.71} & \underline{62.43} & 11.68 & 5.56 & 29.80 & 31.04 \\
      & X-Token (P-KL)$^\dagger$ & \underline{45.63} & 62.19 & 15.69 & 7.78 & \underline{31.00} & \underline{32.46} \\
      & \textbf{\method (ours)} & 45.22 & 61.80 & \textbf{21.30} & \underline{8.26} & \textbf{32.60} & \textbf{33.84} \\
    \midrule
    \multirow{5}{*}{\makecell[l]{Cross-tokenizer\\Phi-4-mini}}
      & ULD & 41.64 & \textbf{63.38} & 19.94 & 6.54 & \underline{30.80} & 32.46 \\
      & ALM & 44.41 & 62.35 & 18.88 & 8.68 & 28.40 & 32.54 \\
      & GOLD & 46.42 & 62.98  & \underline{26.46} & \underline{9.34} & 29.80 & 35.00 \\
      & X-Token (H-KL)$^\dagger$  & \textbf{46.65} & \underline{63.22} & 26.08 & \textbf{9.36} & 30.60 & \underline{35.18} \\
      & \textbf{\method (ours)} & \underline{46.47} & 62.98 & \textbf{26.91} & 9.22 & \textbf{33.60} & \textbf{35.84} \\
    \bottomrule
  \end{tabular}%
  }
\end{table*}

\paragraph{Models and data.}

Our main experiments use Llama-3.2-1B~\citep{grattafiori2024llama}
and Gemma3-1B-PT~\citep{gemma2025gemma3} as students and four
teachers: Llama-3.2-3B and Gemma3-4B-PT are tokenizer-matched
teachers for their respective students, while Qwen3-4B-Base
~\citep{yang2025qwen3} and Phi-4-mini~\citep{abouelenin2025phi}
are cross-tokenizer teachers. To evaluate larger students at different training stages, we also use Llama-3.1-8B: we distill the base
checkpoint from Qwen3-14B-Base and Phi-4-14B, and the
instruction-tuned checkpoint from Qwen3-14B. We use the NVIDIA CLIMB
corpus~\citep{diao2025nemotron}
\footnote{\url{https://huggingface.co/datasets/nvidia/Nemotron-ClimbMix}}
for all base-student experiments and Nemotron-Cascade-SFT
~\citep{yang2026nemotron}
\footnote{\url{https://huggingface.co/datasets/nvidia/Nemotron-Cascade-2-SFT-Data}}
for the instruction-tuned experiment.
\vspace{-0.2in}
\paragraph{Training.}
The 1B experiments run for
$30{,}000$ steps with an effective batch size of $768$
and a context length of $4096$. We use AdamW with $\beta_1=0.9$, $\beta_2=0.95$, gradient clipping at
$1.0$, an initial student learning rate of $5\times10^{-5}$, and a weight decay of $0.001$.
The 8B scaling experiments use the same settings and run
for $2{,}000$ steps. The default \method configuration uses
$\Div=\operatorname{JSD}$ and $\beta=5$.
\vspace{-0.2in}
\paragraph{Evaluation.}
Our primary evaluation for base models comprises MMLU~\citep{hendrycks2020mmlu},
GSM8K~\citep{cobbe2021gsm}, MATH~\citep{hendrycks2021math},
WinoGrande~\citep{sakaguchi2021winogrande}, and MBPP~\citep{austin2021mbpp}. We use 3-shot
prompting, greedy decoding for generative tasks, and the same
lm-evaluation-harness configuration for every method.
For the instruction-tuned student, we evaluate 0-shot GSM8K, MATH,
IFEval~\citep{zhou2023instruction}, MBPP, and
MBPP+~\citep{liu2023your}, following task-specific
instruction-tuned protocols held fixed across methods.
\vspace{-0.2in}
\paragraph{Baselines.}
We compare against the unmodified base model, continued pretraining (CPT)
without a teacher, and same-tokenizer KD. Cross-tokenizer baselines include
ALM~\citep{minixhofer2025universal}, ULD~\citep{boizard2024towards},
GOLD~\citep{patino2025_unlocking_on_policy_distillation_for_any_model_family},
and X-Token~\citep{sreenivas2026x}. For X-Token, P-KL and H-KL denote its
projection and hybrid configurations, respectively. We select P-KL or H-KL for each teacher--student pair following the coverage analysis in the original work. The selected configuration is indicated in each table. For ALM, we use chunk temperature $\tau=10$ and binary
cross-entropy (BCE), without its hidden-state distillation component. Appendix~\ref{app:method-comparison} compares the differences between \method and baselines. In our experiments, all cross-tokenizer methods use the same offset-based alignment, and GOLD, X-Token, and \method use JSD in the main comparisons. For each student setting, all continued-training runs use the same data and update budget. Standalone teacher scores are reported in Table~\ref{tab:teacher-performance} for reference and are excluded from comparisons among student models.

\subsection{Main Results at 1B Scale}
\label{sec:main-results}

\paragraph{Llama-3.2-1B student.}
Table~\ref{tab:llama1b-main} reports the results for Llama-3.2-1B. GOLD can suffer from an \emph{uncommon-token failure}: task-critical
tokens, including numeral tokens important for mathematical reasoning, may
fall outside its matched set and receive harmful supervision, as discussed
by X-Token~\citep{sreenivas2026x}. X-Token handles the two teacher--student
pairs through different loss modes: P-KL with the Qwen teacher and H-KL
with the Phi teacher, both outperforming their corresponding GOLD
baselines. In contrast, \method uses the same event-matching objective
for both pairs and further improves over X-Token by $1.38$ and $0.66$
average points, reaching the highest five-task averages of $33.84$ and
$35.84$, respectively. Across the two teachers, \method also improves
over ULD by $3.38$--$4.66$ points, over ALM by $1.92$--$3.30$ points,
and over GOLD by $0.84$--$2.80$ points. It further exceeds CPT by $3.05$--$5.05$ points and same-tokenizer KD from Llama-3.2-3B by
$2.85$--$4.85$ points. These results show that \method benefits from both cross-tokenizer teachers, outperforming CPT and same-tokenizer KD without pair-specific loss selection.

\begin{table*}[t]
  \caption{Transfer to the Gemma3-1B-PT student. Bold and underlined values denote the best and second-best scores within each cross-tokenizer teacher group. $^\dagger$ X-Token configurations are selected using tokenizer-pair coverage analysis.}
  \label{tab:gemma1b-transfer}
  \centering
  \scriptsize
  \setlength{\tabcolsep}{2pt}
  \renewcommand{\arraystretch}{0.95}
  \resizebox{\linewidth}{!}{%
  \begin{tabular}{ll|*{5}{N}|N}
    \toprule
    \textbf{Setting} & \textbf{Method}
      & \textbf{MMLU} & \textbf{WG}
      & \textbf{GSM8K} & \textbf{MATH} & \textbf{MBPP} & \textbf{Avg.-5} \\
    \midrule
    \multirow{2}{*}{No distillation}
      & Gemma3-1B-PT (base) & 26.44 & 61.72 & 1.59 & 3.58 & 8.20 & 20.31 \\
      & Continued pretraining & 36.00 & 58.96 & 10.77 & 5.10 & 15.00 & 25.17 \\
    \midrule
    Same tokenizer
      & Gemma3-4B-PT & 38.51 & 61.25 & 9.25 & 5.30 & 18.40 & 26.54 \\
    \midrule
    \multirow{5}{*}{\makecell[l]{Cross-tokenizer\\Qwen3-4B}}
      & ULD & 25.99 & \underline{60.85} & 4.47 & 2.96 & 13.20 & 21.49 \\
      & ALM & 37.25 & 60.46 & 11.30 & 5.08 & \underline{17.80} & 26.38 \\
      & GOLD & \textbf{41.23} & 60.69 & 4.32 & 4.62 & 0.00 & 22.17 \\
      & X-Token (P-KL)$^\dagger$ & 39.40 & \textbf{60.93} & \underline{12.59} & \underline{5.10} & 17.20 & \underline{27.04} \\
      & \textbf{\method (ours)} & \underline{39.56} & \textbf{60.93} & \textbf{13.04} & \textbf{5.40} & \textbf{19.60} & \textbf{27.71} \\
    \midrule
    \multirow{5}{*}{\makecell[l]{Cross-tokenizer\\Phi-4-mini}}
       & ULD & 31.63 & 61.07 & 7.88 & 4.08 & 14.00 & 23.73 \\
      & ALM & 38.12 & 61.33 & \underline{12.36} & \underline{5.02} & \underline{19.00} & \underline{27.17} \\
      & GOLD & \underline{41.21} & 60.93 & 1.36 & 3.92 & 0.00 &  21.48 \\
      & X-Token (P-KL)$^\dagger$ & 39.72 & \textbf{62.35} & 1.74 & 3.40 & 10.80 & 23.60 \\
      & \textbf{\method (ours)} & \textbf{41.28} & \underline{61.64} & \textbf{15.62} & \textbf{5.76} & \textbf{20.60} & \textbf{28.98} \\
    \bottomrule
  \end{tabular}%
  }
\end{table*}

\vspace{-0.1in}
\paragraph{Gemma3-1B-PT student.}
Gemma3-1B-PT provides a challenging transfer setting with its
SentencePiece-based tokenizer. As shown in
Table~\ref{tab:gemma1b-transfer}, GOLD yields zero MBPP accuracy with
both teachers and weak mathematical reasoning results, consistent with
the uncommon-token failure discussed above (see Appendix \ref{app:benchmark-token-audit}). As a result, X-Token uses P-KL for both
teacher--student pairs. This configuration is effective with Qwen,
reaching a five-task average of $27.04$, but remains below CPT with Phi:
its average is $23.60$, with GSM8K and MATH scores of only $1.74$ and
$3.40$.

In contrast, \method uses the same common-support objective for both
teachers and achieves the highest five-task averages of $27.71$ with
Qwen and $28.98$ with Phi, exceeding X-Token by $0.67$ and $5.38$
points, respectively. Across the two teachers, \method also improves
over ULD by $5.25$--$6.22$ points and over ALM by $1.33$--$1.81$
points, while outperforming GOLD in both settings. It further exceeds CPT by $2.54$--$3.81$ points and same-tokenizer KD from
Gemma3-4B-PT by $1.17$--$2.44$ points. Across all four 1B teacher--student settings, \method uses the same objective and achieves the highest Avg.-5 among the evaluated methods.

\subsection{Support Selection for Non-One-to-One Chunks}
\label{sec:mismatch-support}

Table~\ref{tab:mismatch-support-ablation} compares three support policies
for non-one-to-one chunks: realized-only support for all such chunks, the
proposed shape-dependent support, and full support for all such chunks.
Realized-only support retains the observed sequence pair, whereas full
support includes all retained admissible final-token pairs, with the observed prefixes fixed. The proposed policy uses full support for one-to-many and many-to-one chunks and realized-only support for many-to-many chunks.
All variants include a residual event. For each teacher, the
one-to-one construction, divergence, $\beta$, student, and training
settings remain unchanged across variants.

\method achieves the highest Avg.-5 with both teachers,
increasing the score from $33.58$ to $33.84$ with Qwen and from
$35.22$ to $35.84$ with Phi compared with realized-only support,
at nearly unchanged step time. These gains suggest that matching
alternative continuations in one-to-many and many-to-one chunks provides useful
additional supervision. Extending full support to many-to-many
chunks yields lower Avg.-5 scores while requiring $2.86\times$ and
$3.92\times$ as much time per step compared with \method.
Thus, expanding the support does not necessarily improve accuracy
in the evaluated settings. These results support using full support
for one-to-many and many-to-one chunks while retaining only the observed pair for many-to-many chunks, balancing distillation quality and computational cost.

\begin{table*}[t]
  \caption{Support-selection ablation for non-one-to-one chunks with
    the Llama-3.2-1B student. Realized-only retains the observed
    sequence pair; full retains all admissible final-token pairs
    with fixed observed prefixes. All variants include a residual
    event. One-to-one support is unchanged across variants. Avg.-5 is the unweighted average of MMLU, WinoGrande, GSM8K, MATH, and MBPP. \textbf{Bold} marks the highest mean for each teacher.}
  \label{tab:mismatch-support-ablation}
  \centering
  \small
  \setlength{\tabcolsep}{5pt}
  \renewcommand{\arraystretch}{1.1}
  \begin{tabular}{lccccc}
    \toprule
    \textbf{Support policy}
      & \textbf{\makecell{$1{:}N$ / $M{:}1$}}
      & \textbf{\makecell{$M{:}N$\\($M,N>1$)}}
      & \textbf{\makecell{Qwen3-4B\\Avg.-5}}
      & \textbf{\makecell{Phi-4-mini\\Avg.-5}}
      & \textbf{\makecell{Avg. time/step (s)\\Qwen / Phi}} \\
    \midrule
    Realized-only for all
      & Realized-only & Realized-only
      & 33.58
      & 35.22
      & 7.45 / 6.99 \\
    \method (proposed)
      & Full & Realized-only
      & \textbf{33.84}
      & \textbf{35.84}
      & 7.45 / 7.06 \\
    Full support for all
      & Full & Full
      & 33.69
      & 35.55
      & 21.31 / 27.64 \\
    \bottomrule
  \end{tabular}
\end{table*}



\subsection{Top-$k$ Robustness and Training Efficiency}
\label{sec:topk-efficiency}

\begin{wrapfigure}[16]{r}{0.5\columnwidth}
  \centering
  \vspace{-0.5in}
  \includegraphics[width=\linewidth]{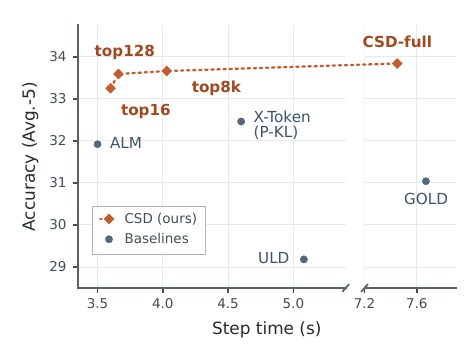}
  \vspace{-0.4in}
  \caption{Accuracy--efficiency trade-off with Qwen3-4B as teacher and
Llama-3.2-1B as student. Lower step time and higher Avg.-5 are better.}
  \label{fig:accuracy-efficiency}
\end{wrapfigure}
Figure~\ref{fig:accuracy-efficiency} compares Avg.-5 and average
training-step time for Qwen3-4B$\rightarrow$Llama-3.2-1B.
We evaluate teacher top-$k$
selection with $k\in\{16,128,8192\}$ alongside full-logit \method
and baselines.

With $k=128$, \method achieves an Avg.-5 of $33.59$, only
$0.25$ points below the full-logit score of $33.84$, while reducing
step time from $7.45$ to $3.66$ seconds, a $2.04\times$ speedup.
Further reducing $k$ to $16$ gives $33.25$ at $3.60$ seconds per
step. With $k=8192$, Avg.-5 reaches $33.66$, only $0.18$ points
below full logits, at $4.03$ seconds per step. The $k=16$ and $k=128$ configurations achieve
higher Avg.-5 with lower step time than ULD, GOLD, and X-Token.
ALM remains slightly faster, but \method with $k=16$ improves
Avg.-5 by $1.33$ points while requiring only $0.10$
additional seconds per step. These
results support a favorable accuracy--efficiency trade-off
under teacher top-$k$ truncation in this setting.

\begin{table*}[t]
  \caption{Scaling results for the Llama-3.1-8B base student. Teacher scores serve as standalone references. Avg.-4 is the unweighted mean over the four displayed tasks.}
  \label{tab:base8b}
  \centering
  \scriptsize
  \setlength{\tabcolsep}{2pt}
  \renewcommand{\arraystretch}{0.95}
  \resizebox{\linewidth}{!}{%
  \begin{tabular}{ll|*{4}{N}|N}
    \toprule
    \textbf{Setting} & \textbf{Method}
      & \textbf{MMLU}
      & \textbf{GSM8K} & \textbf{MATH} & \textbf{MBPP} & \textbf{Avg.-4} \\
    \midrule
    \multirow{2}{*}{No distillation}
      & Llama-3.1-8B (base) & 64.98 &  54.13 & 21.16 & 48.60 & 47.22 \\
      & Continued pretraining & 66.19 &  68.39 & 32.42 & 51.20 & 54.55 \\
    \midrule
    \multirow{2}{*}{\makecell[l]{Teachers\\(reference)}}
      & \textit{Phi-4-14B} & \textit{80.37} &  \textit{91.96} & \textit{50.98} & \textit{72.20} & \textit{73.88} \\
      & \textit{Qwen3-14B-Base} & \textit{80.71}  & \textit{86.73} & \textit{54.84} & \textit{74.20} & \textit{74.12} \\
    \midrule
    \multirow{4}{*}{Cross-tokenizer}
      & Qwen3-14B-Base, ALM & 65.91 & 69.45 & 29.08 & 52.00 & 54.11 \\
      & Qwen3-14B-Base, \method & \textbf{66.35} & \textbf{71.19} & \textbf{35.74} & \textbf{53.60} & \textbf{56.72} \\
       \cmidrule(ll){2-7}
      & Phi-4-14B, ALM & 66.12 & 72.48 & 29.58 & 50.80 & 54.75 \\
      & Phi-4-14B, \method & \textbf{67.74} & \textbf{75.59} & \textbf{37.00} & \textbf{55.60} & \textbf{58.98} \\
    \bottomrule
  \end{tabular}%
  }
\end{table*}
\vspace{-0.1in}
\subsection{Scaling to Larger and Instruction-Tuned Students}
\label{sec:scaling}

To evaluate whether the gains extend beyond 1B students, we study
both Llama-3.1-8B base and instruction-tuned checkpoints. Given the higher training cost, all 8B runs use a budget of $2{,}000$ steps.
\vspace{-0.2in}
\paragraph{Base student.}
We distill the Llama-3.1-8B base student from Qwen3-14B-Base and
Phi-4-14B. Table~\ref{tab:base8b} shows that \method improves over
both CPT and ALM on all four tasks with either
teacher. With Qwen, \method reaches an Avg.-4 of $56.72$, exceeding
CPT and ALM by $2.17$ and $2.61$ points, respectively. With Phi,
it reaches $58.98$, with corresponding gains of $4.43$ and $4.23$
points. The largest gains over ALM occur on MATH: $6.66$ points
with Qwen and $7.42$ points with Phi. These gains extend to the larger base student within the evaluated training budget.

\begin{table*}[t]
  \caption{Results for the Llama-3.1-8B-Instruct student.  IFEval-p and
    IFEval-i denote prompt- and instruction-level strict accuracy.  Evaluation
    follows each task's standard instruction-tuned protocol.  Avg.-6 is the
    unweighted mean over the six displayed metric columns.}
  \label{tab:instruct8b}
  \centering
  \scriptsize
  \setlength{\tabcolsep}{2pt}
  \renewcommand{\arraystretch}{1}
  \resizebox{\linewidth}{!}{%
  \begin{tabular}{ll|*{6}{N}|N}
    \toprule
    \textbf{Setting} & \textbf{Method}
      & \textbf{GSM8K} & \textbf{MATH} & \textbf{IFEval-p}
      & \textbf{IFEval-i} & \textbf{MBPP} & \textbf{MBPP+} & \textbf{Avg.-6} \\
    \midrule
    \multirow{2}{*}{No distillation}
      & Llama-3.1-8B-Inst & 75.36 & 46.18 & 74.49 & 82.01 & 64.02 & 52.65 & 65.79 \\
      & Continued training & 72.48 & 40.72 & 71.72 & 78.30 & 64.20 & 55.90 & 63.89 \\
    \midrule
    Teacher (reference)
      & \textit{Qwen3-14B} & \textit{78.62} & \textit{78.76} & \textit{84.47} & \textit{89.57} & \textit{83.79} & \textit{72.34} & \textit{81.26} \\
    \midrule
    \multirow{2}{*}{Cross-tokenizer}
      & Qwen3-14B, ALM & \textbf{79.76} & 46.72 & 70.06 & 78.06 & 65.66 & \textbf{56.89} & 66.19 \\
      & Qwen3-14B, \method & 76.50 & \textbf{47.76} & \textbf{76.71} & \textbf{83.45} & \textbf{65.91} & 56.39 & \textbf{67.79} \\
    \bottomrule
  \end{tabular}%
  }
\end{table*}

\vspace{-0.2in}
\paragraph{Instruction-tuned student.}
We further evaluate \method with Llama-3.1-8B-Instruct as the student and Qwen3-14B as the teacher. Table~\ref{tab:instruct8b} shows that \method achieves the highest student Avg.-6 of $67.79$, outperforming the initial checkpoint, continued training, and ALM by $2.00$, $3.90$, and $1.60$ points, respectively. It exceeds ALM on four of six metrics, including gains of $6.65$ and $5.39$ points on IFEval-p and IFEval-i. Together, the base and instruction-tuned results show that
\method improves average performance over teacher-free continued training
and ALM for 8B students, extending the gains observed at 1B scale to
larger students at different training stages.


\vspace{-0.1in}
\section{Conclusion}
\label{sec:conclusion}

We introduced \method, a cross-tokenizer KD method that matches normalized teacher and student distributions over a shared support of common-token and multi-token events, with a residual event capturing the remaining probability mass. Its unified objective applies across chunk shapes and tokenizer pairs without pair-specific loss selection, matching teacher probabilities over alternative events where available.

\method achieves the highest average score among the evaluated
baselines across the four 1B teacher--student settings and
outperforms continued training and ALM for both base and
instruction-tuned 8B students. For the instruction-tuned student,
it also improves all six reported metrics over the initial checkpoint,
including instruction-following accuracy. The ablations support using full support for one-to-many and many-to-one chunks
and realized-only support for many-to-many chunks, which achieves the highest mean accuracy without
the additional cost of expanding many-to-many support.
In the Qwen–Llama setting, teacher top-$k$ truncation substantially reduces training-step time with only small changes in accuracy.

Our evaluation focuses primarily on English tasks, and the residual event provides aggregate supervision
without distinguishing the teacher's preferences among outcomes
outside the explicit support.
Future work will explore more efficient construction of richer common
support, particularly for many-to-many chunks, task-specific
choices of $\beta$, and generalization across languages and tokenizer families.


\clearpage
{
  \small
  \bibliographystyle{unsrt}
  \bibliography{paper}
}

\appendix

\section{Tokenizer and Alignment Statistics}
\label{app:stats}

We characterize vocabulary overlap, task-relevant mapping gaps, and aligned-chunk statistics across the four teacher--student pairs with 1B students. Pair labels use teacher$\rightarrow$student order, whereas chunk shapes use student:teacher token counts.

\subsection{Tokenizer Overlap}

Table~\ref{tab:tokenizer-overlap} summarizes the common-token map for
each pair. The common-pair count, $|\MatchSet|$, is the number of
retained one-to-one token correspondences used to construct explicit
events for one-to-one chunks. Because each token appears in at most
one pair, this count also gives the number of covered tokens on
each side. Student and teacher coverage divide this count by their
respective text-distillation vocabulary sizes.

\begin{table}[ht]
  \caption{Vocabulary overlap for the evaluated teacher--student pairs.
    Common pairs are retained one-to-one token correspondences.
    Student and teacher coverage divide the common-pair count by the
    respective text-distillation vocabulary size.}
  \label{tab:tokenizer-overlap}
  \centering
  \small
  \setlength{\tabcolsep}{5pt}
  \renewcommand{\arraystretch}{1.1}
  \resizebox{\linewidth}{!}{%
  \begin{tabular}{lccccc}
    \toprule
    \textbf{Teacher $\rightarrow$ student}
      & \textbf{\makecell{Student vocab.\\size}}
      & \textbf{\makecell{Teacher vocab.\\size}}
      & \textbf{\makecell{Common\\pairs}}
      & \textbf{\makecell{Student\\covered}}
      & \textbf{\makecell{Teacher\\covered}} \\
    \midrule
    Qwen3-4B $\rightarrow$ Llama-3.2-1B
      & 128,256 & 151,669 & 108,581 & 84.66\% & 71.59\% \\
    Phi-4-mini $\rightarrow$ Llama-3.2-1B
      & 128,256 & 200,029 & 102,149 & 79.64\% & 51.07\% \\
    \midrule
    Qwen3-4B $\rightarrow$ Gemma3-1B-PT
      & 262,144 & 151,669 & 94,932 & 36.21\% & 62.59\% \\
    Phi-4-mini $\rightarrow$ Gemma3-1B-PT
      & 262,144 & 200,029 & 130,704 & 49.86\% & 65.34\% \\
    \bottomrule
  \end{tabular}%
  }
\end{table}

Common pairs cover $79.64\%$--$84.66\%$ of the Llama student vocabulary, compared with $36.21\%$--$49.86\%$ for Gemma. These statistics measure vocabulary-entry coverage, not probability mass or aligned-chunk frequency. Unpaired tokens contribute through
the residual event in the one-to-one construction, while virtual support for non-one-to-one chunks is constructed separately. Despite differences in vocabulary overlap, \method uses the same distribution-matching objective for all four pairs.

\subsection{Task-Relevant Tokens Outside the Common Map}
\label{app:benchmark-token-audit}

Tables~\ref{tab:gsm8k-uncommon-examples}
and~\ref{tab:mbpp-uncommon-examples} list the three most frequent
entries in each pair's exported audit example list. 
\emph{Occurrences} counts token appearances, while \emph{Examples}
counts distinct examples containing the token. GSM8K references
include reasoning and final answers; MBPP references are programs
with line endings normalized to LF. Spaces, tabs, and newlines are
shown as \texttt{\textvisiblespace{}}, \texttt{\textbackslash{}t},
and \texttt{\textbackslash{}n}, respectively.
\vspace{-0.2in}
\paragraph{GSM8K.}
The audited examples include numeral tokens for
Qwen--Llama and Phi--Gemma,
and unit or rate fragments for the other pairs. Losing token-specific supervision
for numerals can weaken supervision of the quantities used in
intermediate calculations and final answers. The numeral mapping
gaps therefore suggest a plausible explanation for GOLD's low
GSM8K and MATH scores in the Qwen--Llama and Phi--Gemma settings
(Tables~\ref{tab:llama1b-main} and~\ref{tab:gemma1b-transfer}).
\vspace{-0.2in}
\paragraph{MBPP.}
For pairs with Gemma students, the audited unmatched tokens
include frequent whitespace and code-syntax fragments, whereas
the Llama examples include numerals and identifier fragments.
The whitespace and syntax fragments encode indentation, line
breaks, and block boundaries needed for valid Python programs.
Inadequate supervision of these elements can make otherwise
correct solutions syntactically invalid. These frequent mapping
gaps therefore suggest a plausible explanation for GOLD's zero
MBPP accuracy with both Gemma teacher--student pairs (Table~\ref{tab:gemma1b-transfer}): losing token-specific supervision for code structure can undermine programming performance.



\begin{table*}[t]
  \centering
  \begin{minipage}[t]{0.49\textwidth}
    \centering
    \caption{Uncommon-token examples in $1{,}319$ GSM8K reference solutions.}
    \label{tab:gsm8k-uncommon-examples}
    \small
    \setlength{\tabcolsep}{4pt}
    \renewcommand{\arraystretch}{1.05}
    \begin{tabular*}{\linewidth}{@{\extracolsep{\fill}}ccc@{}}
      \toprule
      \textbf{Token} & \textbf{Occurrences} & \textbf{Examples} \\
      \midrule
      \multicolumn{3}{l}{\textbf{Qwen3-4B $\rightarrow$ Llama-3.2-1B}} \\
      \texttt{10} & 1,228 & 359 \\
      \texttt{20} & 1,222 & 334 \\
      \texttt{12} & 909 & 256 \\
      \midrule
      \multicolumn{3}{l}{\textbf{Phi-4-mini $\rightarrow$ Llama-3.2-1B}} \\
      \texttt{/week} & 40 & 14 \\
      \texttt{/hour} & 28 & 17 \\
      \texttt{Subtract} & 14 & 13 \\
      \midrule
      \multicolumn{3}{l}{\textbf{Qwen3-4B $\rightarrow$ Gemma3-1B-PT}} \\
      \texttt{/day} & 55 & 21 \\
      \texttt{/h} & 29 & 18 \\
      \texttt{\%\textbackslash{}n} & 24 & 18 \\
      \midrule
      \multicolumn{3}{l}{\textbf{Phi-4-mini $\rightarrow$ Gemma3-1B-PT}} \\
      \texttt{10} & 1,228 & 359 \\
      \texttt{20} & 1,222 & 334 \\
      \texttt{12} & 909 & 256 \\
      \bottomrule
    \end{tabular*}
  \end{minipage}
  \hfill
  \begin{minipage}[t]{0.49\textwidth}
    \centering
    \caption{Uncommon-token examples in $500$ MBPP reference programs.}
    \label{tab:mbpp-uncommon-examples}
    \small
    \setlength{\tabcolsep}{4pt}
    \renewcommand{\arraystretch}{1.05}
    \begin{tabular*}{\linewidth}{@{\extracolsep{\fill}}ccc@{}}
      \toprule
      \textbf{Token} & \textbf{Occurrences} & \textbf{Examples} \\
      \midrule
      \multicolumn{3}{l}{\textbf{Qwen3-4B $\rightarrow$ Llama-3.2-1B}} \\
      \texttt{10} & 35 & 18 \\
      \texttt{100} & 7 & 6 \\
      \texttt{26} & 5 & 3 \\
      \midrule
      \multicolumn{3}{l}{\textbf{Phi-4-mini $\rightarrow$ Llama-3.2-1B}} \\
      \texttt{\textvisiblespace{}heapq} & 21 & 9 \\
      \texttt{rlen} & 11 & 1 \\
      \texttt{\_odd} & 10 & 6 \\
      \midrule
      \multicolumn{3}{l}{\textbf{Qwen3-4B $\rightarrow$ Gemma3-1B-PT}} \\
      \texttt{\textvisiblespace{}\textbackslash{}n} & 908 & 216 \\
      \texttt{):\textbackslash{}n} & 431 & 314 \\
      \texttt{\textbackslash{}t} & 314 & 71 \\
      \midrule
      \multicolumn{3}{l}{\textbf{Phi-4-mini $\rightarrow$ Gemma3-1B-PT}} \\
      \texttt{\textvisiblespace{}\textbackslash{}n} & 908 & 216 \\
      \texttt{):\textbackslash{}n} & 433 & 314 \\
      \texttt{\textbackslash{}t} & 314 & 71 \\
      \bottomrule
    \end{tabular*}
  \end{minipage}
\end{table*}

\vspace{-0.1in}
\subsection{Chunk-Shape Distribution}
\label{app:chunk-distribution}

Table~\ref{tab:chunk-distribution} reports recorded chunk-shape
frequencies. Chunk shapes use student:teacher
token counts. One-to-many chunks have shape $1{:}N$ with $N>1$,
many-to-one chunks have shape $M{:}1$ with $M>1$, and many-to-many
chunks have shape $M{:}N$ with $M,N>1$. Many-to-many chunks need
not contain equal numbers of tokens on the two sides.

\begin{table}[t]
  \caption{Recorded chunk-shape frequencies (\%). The final column
combines all non-one-to-one chunks. Totals may differ slightly
from component sums due to rounding.}
  \label{tab:chunk-distribution}
  \centering
  \small
  \setlength{\tabcolsep}{4pt}
  \begin{tabular}{lccccc}
    \toprule
    \textbf{Teacher $\rightarrow$ student}
      & $\mathbf{1{:}1}$ & $\mathbf{1{:}N}$
      & $\mathbf{M{:}1}$ & $\mathbf{M{:}N}$
      & \textbf{All non-$1{:}1$} \\
    \midrule
    Qwen3-4B $\rightarrow$ Llama-3.2-1B
      & 98.17 & 1.61 & 0.16 & 0.06 & 1.83 \\
    Phi-4-mini $\rightarrow$ Llama-3.2-1B
      & 97.05 & 0.62 & 1.70 & 0.63 & 2.95 \\
    Qwen3-4B $\rightarrow$ Gemma3-1B-PT
      & 90.55 & 3.00 & 4.87 & 1.58 & 9.45 \\
    Phi-4-mini $\rightarrow$ Gemma3-1B-PT
      & 89.75 & 2.49 & 6.32 & 1.44 & 10.25 \\
    \bottomrule
  \end{tabular}
\end{table}

One-to-one chunks dominate for every pair. Non-one-to-one chunks
are nevertheless more frequent with Gemma ($9.45$--$10.25\%$) than
with Llama ($1.83$--$2.95\%$), indicating greater segmentation
differences with the evaluated teachers. One-to-many and many-to-one
chunks account for most non-one-to-one chunks; many-to-many chunks constitute at most $1.58\%$ of all recorded chunks.
\vspace{-0.1in}
\subsection{Common-Support Size Distribution}
\label{app:support-size-distribution}

We measure explicit common-support sizes for one-to-many and
many-to-one chunks using the same $500$ text samples
across all four teacher--student pairs, a context length of $4096$,
and no top-$k$ truncation. For each recorded chunk, $K$ counts
deduplicated explicit events with the observed prefix and
multi-token length fixed. The residual event is excluded, so the
corresponding event distribution has $K+1$ entries.

\begin{figure}[t]
  \centering
  \includegraphics[width=\linewidth]{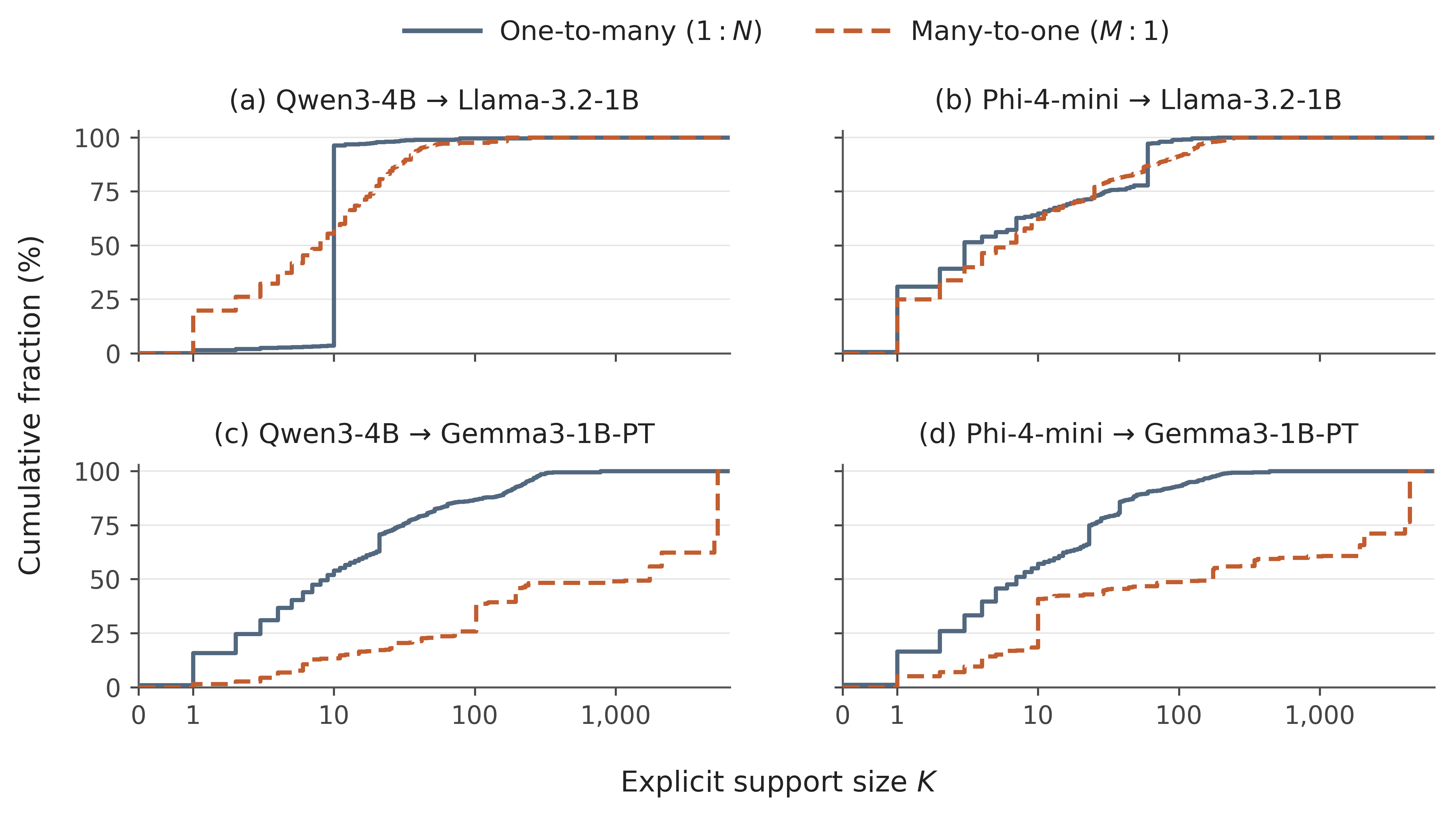}
  \caption{Common-support size distributions for one-to-many and many-to-one chunks. Each panel compares one-to-many and many-to-one chunks for a teacher--student pair. Curves show the percentage of recorded chunks with at most $K$ explicit events, excluding the residual event. The horizontal axis is linear from $0$ to $1$ and logarithmic above $1$.}
  \label{fig:support-size-distribution}
\end{figure}

Figure~\ref{fig:support-size-distribution} shows empirical cumulative distributions, normalized separately for each pair and chunk shape. Each recorded chunk occurrence has equal weight. The curves include all recorded chunks, including those with no explicit event ($K=0$), which do not contribute to KD. Chunks with $K=1$ are accepted and form a binary distribution consisting of one explicit event and the residual.

One-to-many supports are typically small, with median sizes ranging from $3$ to $10$ across the four pairs. Many-to-one supports vary more substantially: the median sizes are $8$ and $6$ for the Llama student with Qwen and Phi teachers, respectively, but increase to $1{,}740$ and $175$ for the corresponding Gemma pairs. The largest recorded supports also occur in the Gemma many-to-one setting, reaching $5{,}294$ events with Qwen and $4{,}362$ with Phi. Thus, although the same common-support construction applies across pairs, the number of explicit events depends strongly on both the tokenizer pair and the direction of the segmentation mismatch.

\section{Offset-Based Sequence Alignment}
\label{app:alignment}

\subsection{Inputs and outputs}

Our alignment procedure uses the
character spans returned by the student and teacher tokenizers for the
same text $x$. Following Section~\ref{sec:aligned-spans}, these spans are
$[a_r^S,b_r^S)$ for student tokens and $[a_r^T,b_r^T)$ for teacher tokens.
Offsets refer to character positions in the original text, rather than
byte positions or lengths of decoded prefixes. We retain the original
token indices to associate each chunk with its model predictions.

The procedure returns a set of valid text chunks $\ChunkSet$. Each
chunk records its starting indices $(i,j)$, the indices immediately
after it $(i+M,j+N)$, and its shared source interval $[a_c,b_c)$.
These records identify the token ranges, their shape $M{:}N$, and
their preceding histories
$h_S^c=\mathbf{s}_{<i}$ and $h_T^c=\mathbf{t}_{<j}$.
Unmatched ranges and candidates that fail verification are marked
invalid. 

We scan both token sequences in order, maintaining one pointer per
sequence:
\begin{enumerate}[leftmargin=*]
  \item \textbf{Locate a shared start.}
        A candidate begins when the current student and teacher
        tokens have the same start offset. Denoting their indices
        by $i$ and $j$, we set $a_c=a_i^S=a_j^T$.
        Ranges without corresponding source coverage are marked
        as \emph{orphans}, and the scan continues to the next
        matchable ranges.

  \item \textbf{Initialize the ranges.}
        Include one token from each side, setting $M=N=1$.
        The current end offsets are $b_{i+M-1}^S$ and
        $b_{j+N-1}^T$.

  \item \textbf{Extend the earlier-ending range.}
        If $b_{i+M-1}^S<b_{j+N-1}^T$, include the next student
        token and increment $M$. Otherwise, if
        $b_{j+N-1}^T<b_{i+M-1}^S$, include the next teacher
        token and increment $N$. Compare the end offsets after
        each extension; the side being extended may change
        as the scan proceeds.

  \item \textbf{Close the candidate.}
        Stop at the first coincident end offset for which both
        ranges cover the same source interval, and set
        $b_c=b_{i+M-1}^S=b_{j+N-1}^T$.
        The index pairs $(i,j)$ and $(i+M,j+N)$ delimit the
        candidate. If matching coverage cannot be established,
        mark the unmatched range invalid.

  \item \textbf{Verify the decoded text.}
        Apply the checks below to the candidate token sequences
        $\mathbf{s}^c$ and $\mathbf{t}^c$. Add the chunk to
        $\ChunkSet$ only if verification succeeds.

  \item \textbf{Continue the scan.}
        Resume after the processed ranges. Pointers never move
        backward, and processed tokens are not reconsidered.
        Any trailing range without a counterpart is marked
        as an orphan.
\end{enumerate}

Every candidate passed to verification satisfies
\begin{equation}
  a_i^S=a_j^T=a_c,
  \qquad
  b_{i+M-1}^S=b_{j+N-1}^T=b_c,
  \qquad a_c<b_c.
  \label{eq:common-boundary}
\end{equation}
The intervening token sequences are $\mathbf{s}^c$ and
$\mathbf{t}^c$, as defined in Section~\ref{sec:aligned-spans}.
Their lengths determine whether the chunk is one-to-one,
one-to-many, many-to-one, or many-to-many; no shape is prescribed
in advance. Extending to a shared end offset accommodates different
segmentations without requiring correspondence between internal
token positions.

Invalid ranges are excluded from aligned KD, but their tokens
remain in the original sequences and in subsequent conditioning
histories.

\vspace{-0.1in}
\subsection{Decoded-text verification}

Matching source coverage identifies a candidate but does not
automatically validate it. We verify each complete candidate range
rather than repeatedly decoding the preceding histories. Each range
is decoded with its own tokenizer and normalized using Unicode NFC:
\[
  d_S^c=\operatorname{NFC}\!\left(\Dec_S(\mathbf{s}^c)\right),
  \qquad
  d_T^c=\operatorname{NFC}\!\left(\Dec_T(\mathbf{t}^c)\right).
\]
The comparison uses decoded text, not vocabulary token spellings.
Decoding the complete range also avoids requiring individual tokens
to represent complete text fragments.

A candidate passes verification if any of the following conditions
holds:
\begin{itemize}[leftmargin=*]
  \item \textbf{Exact match:} $d_S^c=d_T^c$.
  \item \textbf{Whitespace-stripped match:} the whitespace-stripped
        forms of $d_S^c$ and $d_T^c$ are identical.
  \item \textbf{Pure-whitespace match:} both decoded strings consist
        entirely of whitespace.
\end{itemize}
These checks accommodate tokenizer-specific whitespace and decoding
artifacts. Candidates that fail all three checks are marked invalid.

A valid text chunk therefore requires both matching source coverage
and an accepted decoded-text comparison. NFC normalization and
whitespace equivalence are used only for verification: they do not
modify the original offsets, token ranges, or histories used to
compute model probabilities.

\begin{algorithm}[t]
\caption{Common-Support Distillation: one training batch}
\label{alg:csd-training}
\footnotesize
\begin{algorithmic}[1]
\Require Text batch $\mathcal{X}$; student $S$ and frozen teacher $T$;
         maps $\mathcal{M},\mathcal{B}$
\Require Divergence $\mathcal{D}$;
         non-one-to-one weight $\beta\geq 0$
\Ensure Joint training loss $\mathcal{L}_{\mathrm{train}}$
\State Tokenize $\mathcal{X}$ with both tokenizers, retaining character offsets
\State Compute full-vocabulary probabilities $p_{S,r}$ and $p_{T,r}$
       using the observed token histories
\State $\mathcal{L}_{\mathrm{CE}}\gets$ student next-token cross-entropy on the batch
\State $L_{\Sigma}\gets 0$; $n_{\mathrm{acc}}\gets 0$
\For{each text $x\in\mathcal{X}$ with token sequences $\mathbf{s},\mathbf{t}$}
  \State $\mathcal{C}(x)\gets
         \Call{AlignOffsets}{\mathbf{s},\mathbf{t},\text{offsets}}$
  \For{each chunk $c=(i,j,M,N)\in\mathcal{C}(x)$}
    \State $\mathbf{s}^c\gets\mathbf{s}_{i:i+M-1}$;
           $\mathbf{t}^c\gets\mathbf{t}_{j:j+N-1}$
    \If{$M=N=1$} \Comment{One-to-one chunks}
      \State $\mathcal{A}_c\gets\mathcal{M}$
    \ElsIf{$\min(M,N)=1$} \Comment{One-to-many or many-to-one chunks}
      \State $\mathcal{A}_c\gets
        \Call{PairedCompletions}{\mathcal{B},\mathbf{s}_{<M}^c,
                                \mathbf{t}_{<N}^c,M,N}$
    \Else \Comment{Many-to-many chunks}
      \State $\mathcal{A}_c\gets\{(s_M^c,t_N^c)\}$ if
        $\operatorname{Dec}_S(\mathbf{s}^c)=\operatorname{Dec}_T(\mathbf{t}^c)$;
        otherwise $\mathcal{A}_c\gets\varnothing$
    \EndIf
    \If{$\mathcal{A}_c=\varnothing$} \Comment{Skip chunks with no explicit event}
      \State \textbf{continue}
    \EndIf
    \State Enumerate $\mathcal{A}_c=\{(u_a,v_a)\}_{a=1}^{K_c}$
    \If{$M=N=1$} \Comment{Direct next-token probabilities}
      \State $q_S^c(a)\gets p_{S,i}(u_a)$;
             $q_T^c(a)\gets p_{T,j}(v_a)$,
             for $a=1,\ldots,K_c$
    \Else \Comment{Chain-rule probabilities}
      \State $\alpha_S\gets
        \prod_{r=1}^{M-1}p_{S,i+r-1}(s_r^c)$;
        $\alpha_T\gets\prod_{r=1}^{N-1}p_{T,j+r-1}(t_r^c)$
      \State $q_S^c(a)\gets\alpha_S p_{S,i+M-1}(u_a)$;
             $q_T^c(a)\gets\alpha_T p_{T,j+N-1}(v_a)$,
             for $a=1,\ldots,K_c$
    \EndIf
    \Statex \Comment{Collect all remaining mass in a residual event}
    \State $P_m^c\gets\bigl(q_m^c(1),\ldots,q_m^c(K_c),
                 1-\sum_{a=1}^{K_c}q_m^c(a)\bigr)$,
           for $m\in\{S,T\}$
    \State $w_c\gets 1$ if $M=N=1$; otherwise $w_c\gets\beta$
    \State $L_{\Sigma}\gets L_{\Sigma}+w_c\mathcal{D}(P_T^c,P_S^c)$;
           $n_{\mathrm{acc}}\gets n_{\mathrm{acc}}+1$
  \EndFor
\EndFor
\Statex \Comment{Average over accepted chunks, with an unweighted count}
\State $\mathcal{L}_{\mathrm{KD}}\gets
        L_{\Sigma}/\max(1,n_{\mathrm{acc}})$
\If{$\mathcal{L}_{\mathrm{KD}}>0$} \Comment{Dynamically balance KD and cross-entropy}
  \State $\mathcal{L}_{\mathrm{train}}\gets
    \operatorname{sg}\!\left(
       \mathcal{L}_{\mathrm{CE}}/\mathcal{L}_{\mathrm{KD}}\right)
    \mathcal{L}_{\mathrm{KD}}+\mathcal{L}_{\mathrm{CE}}$
\Else
  \State $\mathcal{L}_{\mathrm{train}}\gets\mathcal{L}_{\mathrm{CE}}$
\EndIf
\State \Return $\mathcal{L}_{\mathrm{train}}$
\end{algorithmic}
\end{algorithm}
\vspace{-0.2in}

\subsection{Special tokens and the KD interface}

BOS, EOS, and padding tokens are matched separately by semantic role,
rather than by character spans. EOS and padding may be treated as
equivalent when permitted by the tokenizer-role configuration;
their token IDs need not coincide across models.

Alignment supplies the valid observed token ranges, their histories,
and their chunk shapes. Common-support construction then determines
the explicit events and their probabilities, as described in
Section~\ref{sec:common-support}. The whitespace allowances used
during alignment do not replace the text-matching conditions used
to construct that support. A valid aligned chunk with no retained
explicit event contributes no KD term, as defined in
Section~\ref{sec:divergence}.




\section{CSD Training Algorithm}
\label{app:csd-pseudocode}

Algorithm~\ref{alg:csd-training} summarizes the computation of the CSD
training objective for one batch under the default support policy, without teacher
top-$k$ truncation. Teacher top-$k$ selection is described separately
in Section~\ref{sec:topk-support}.

\begin{table*}[t]
\caption{Comparison of \method with the published formulations
of the cross-tokenizer distillation baselines evaluated in this
work. Pair-specific loss choice refers to selecting between loss
modes, not constructing tokenizer-dependent mappings.}
\label{tab:method-comparison}
\centering
\scriptsize
\setlength{\tabcolsep}{3pt}
\renewcommand{\arraystretch}{1.15}
\begin{tabularx}{\linewidth}{@{}
  >{\raggedright\arraybackslash}p{0.15\linewidth}
  >{\raggedright\arraybackslash}X
  >{\raggedright\arraybackslash}X
  >{\centering\arraybackslash}p{0.09\linewidth}
  >{\centering\arraybackslash}p{0.11\linewidth}@{}}
\toprule
\textbf{Method}
& \textbf{Matching rule}
& \textbf{Handling non-one-to-one chunks}
& \textbf{Residual event}
& \textbf{Pair-specific loss choice} \\
\midrule
ULD~\citep{boizard2024towards}
& $L_1$ between sorted, zero-padded probability vectors
& No chunk construction; index-wise comparison up to the shorter sequence
& No & No \\
\addlinespace
GOLD~\citep{patino2025_unlocking_on_policy_distillation_for_any_model_family}
& KL on exact token-string matches + sorted $L_1$ on unmatched entries
& Scales the first-position probability vector by multiplying subsequent realized token probabilities
& No & No \\
\addlinespace
X-Token~\citep{sreenivas2026x}
& P-KL: vocabulary projection + KL; \;\;\;H-KL: KL on relaxed token matching + sorted $L_1$ on unmatched entries
& Scales the first-position probability vector by multiplying subsequent realized token probabilities
& No & Yes \\
\addlinespace
ALM~\citep{minixhofer2025universal}
& Binary divergence on the observed event and a residual
& Chain-rule observed-only probability
& Yes & No \\
\midrule
\textbf{\method (ours)}
& Divergence between distributions over explicit paired events
  and a residual
& Chain-rule common-support probabilities 
& Yes & No \\
\bottomrule
\end{tabularx}

\vspace{0.5em}
\begin{minipage}{\linewidth}
\scriptsize
\textit{Notes.} A residual explicitly collects mass outside individually
matched events; its absence does not imply unnormalized distributions.
X-Token selects P-KL or H-KL through a tokenizer-pair coverage audit. Our evaluated ULD baseline additionally uses the
same chunk-merging procedure as GOLD before applying the ULD loss
\end{minipage}
\end{table*}

\paragraph{Alignment.}
\textsc{AlignOffsets} follows Appendix~\ref{app:alignment}:
it identifies shared character boundaries and retains chunks
that pass the decoded-text checks. 
\vspace{-0.2in}
\paragraph{Precomputed support index.}
We construct vocabulary-conversion tables in both directions.
For each ordinary vocabulary token, we first seek a canonicalized
token-surface match; otherwise, we decode the token and retokenize
the result with the other tokenizer without adding special tokens.
Special tokens are matched by semantic role, including EOS.
Each direction stores one deterministic conversion per
successfully mapped source token.

For a stored sequence $(u_1,\ldots,u_n)$, the index key consists
of its length $n$ and token-ID prefix $(u_1,\ldots,u_{n-1})$.
The associated entry pairs the source token with $u_n$.
For one-to-many chunks, \textsc{PairedCompletions} queries
the student-to-teacher index using the observed teacher
prefix and token count. For many-to-one chunks, it queries
the reverse index using the observed student prefix and
token count. Within each bucket, we process source tokens
in increasing ID order and retain pairs without repeated
token IDs on either side. The indices are built once per
process and cached.

Here, full support refers to all retained candidates in
the corresponding index bucket, rather than an enumeration
of every possible tokenization. Given a constructed prefix
key, bucket retrieval takes expected $O(1)$ time.
Constructing the key depends on prefix length, while
gathering and scoring $K$ candidates requires $O(K)$ work
in addition to prefix scoring.
\vspace{-0.2in}
\paragraph{Probability computation and optimization.}
Empty prefix products equal one, so the same probability computation
also applies to one-to-one chunks. The residual contains all mass
outside the retained support. The denominator counts accepted chunks
without weighting by $\beta$. During backpropagation,
$\operatorname{sg}$ stops gradients through the entire scaling ratio,
while student gradients flow through all retained chain factors
and residual probabilities; the teacher remains fixed.

\section{Comparison with Logits-Based Distillation Methods}
\label{app:method-comparison}

Table~\ref{tab:method-comparison} compares \method with the evaluated logits-based
baselines, distinguishing the distribution-matching rule from
the construction of supervision for non-one-to-one chunks.
The table follows published formulations; experimental adaptations
are described separately in Section~\ref{sec:experimental-setup}.

\begin{table*}[t]
  \caption{Standalone teacher performance under the same 3-shot
    evaluation protocol used for the 1B students.
    Avg.-5 is the unweighted mean over the five displayed tasks.}
  \label{tab:teacher-performance}
  \centering
  \small
  \setlength{\tabcolsep}{5pt}
  \renewcommand{\arraystretch}{1.1}
  \begin{tabular}{lrrrrrr}
    \toprule
    \textbf{Teacher}
      & \textbf{MMLU} & \textbf{WG}
      & \textbf{GSM8K} & \textbf{MATH}
      & \textbf{MBPP} & \textbf{Avg.-5} \\
    \midrule
    Llama-3.2-3B
      & 55.97 & 70.01 & 24.94 & 8.82 & 38.20 & 39.59 \\
    Gemma3-4B-PT
      & 59.34 & 73.16 & 35.78 & 15.26 & 45.80 & 45.87 \\
    Qwen3-4B-Base
      & 72.43 & 72.30 & 84.61 & 27.76 & 67.60 & 64.94 \\
    Phi-4-mini
      & 68.32 & 74.90 & 82.71 & 19.30 & 56.00 & 60.25 \\
    \bottomrule
  \end{tabular}
\end{table*}

\begin{table*}[t]
  \caption{Divergence and chunk-weight ablations with Qwen3-4B and
    Llama-3.2-1B. Divergence comparisons fix $\beta=5$; weight
    comparisons use JSD. FKL and RKL denote forward and reverse KL.
    All other settings are fixed. $\dagger$ marks the shared default,
    reported once. Avg.-5 averages the five displayed tasks.
    \textbf{Bold} marks the highest reported average.}
  \label{tab:objective-ablation}
  \centering
  \scriptsize
  \setlength{\tabcolsep}{2pt}
  \renewcommand{\arraystretch}{0.95}
  \resizebox{0.95\linewidth}{!}{%
  \begin{tabular}{lcc*{6}{N}}
    \toprule
    \textbf{Teacher} & \textbf{Divergence} & \textbf{$\beta$}
      & \textbf{MMLU} & \textbf{WG} & \textbf{GSM8K}
      & \textbf{MATH} & \textbf{MBPP} & \textbf{Avg.-5} \\
    \midrule
    \multirow{5}{*}{Qwen3-4B}
      & JSD & $5^{\dagger}$
      & 45.22 & 61.80 & 21.30 & 8.26 & 32.60 & 33.84 \\
      & FKL & $5$
      & 44.55 & 62.59 & 19.71 & 8.60 & 27.80 & 32.65 \\
      & RKL & $5$
      & 45.48 & 62.43 & 23.05  & 7.30 & 30.60 & 33.77 \\
    \cmidrule(lr){2-9}
      & JSD & $1$
      & 45.24 & 62.35 & 19.71 & 7.12 & 30.00 & 32.88 \\
      & JSD & $20$
      & 45.03 & 62.51 & 23.20 & 8.82 & 30.00 & \textbf{33.91} \\
    \bottomrule
  \end{tabular}%
  }
\end{table*}

Like ALM, \method includes a residual event without pair-specific
loss-mode selection.
The difference lies in the explicit support:
ALM matches the observed aligned chunk and its complement, whereas
\method can match multiple alternative events for one-to-one,
one-to-many, and many-to-one chunks. Under its default support
policy, \method retains binary support for many-to-many chunks,
consisting of the observed sequence pair and the residual.

\section{More Evaluations}
\label{app:eval}

Table~\ref{tab:teacher-performance} reports standalone scores for
all teachers used in the 1B experiments under the same 3-shot
evaluation protocol. Teacher scores are reported separately and are
not included in the rankings of student methods.

Both cross-tokenizer teachers achieve higher Avg.-5 than the
same-tokenizer teachers, providing an opportunity to transfer
knowledge from stronger models outside the student's family.
Qwen has the highest average ($64.94$) and leads on four tasks,
while Phi ($60.25$) performs best on WinoGrande. This ranking does
not carry over directly to distilled students: with \method,
Phi produces higher student averages than Qwen for both Llama
and Gemma (Tables~\ref{tab:llama1b-main}
and~\ref{tab:gemma1b-transfer}). Standalone teacher accuracy
therefore does not by itself determine transfer effectiveness.

\subsection{Divergence and Chunk Weighting}
Table~\ref{tab:objective-ablation} examines the divergence and
non-one-to-one chunk weight $\beta$ with the Qwen3-4B teacher and
Llama-3.2-1B student. At $\beta=5$, JSD achieves an Avg.-5 of
$33.84$, compared with $32.65$ for FKL and $33.77$ for RKL.
JSD has the highest reported average among the three divergences, with performance close to RKL.
The common-support construction accommodates all three choices.

With JSD, increasing $\beta$ from $1$ to $5$ raises Avg.-5 from
$32.88$ to $33.84$. Increasing it further to $20$ yields $33.91$,
a further small increase. These results support
upweighting non-one-to-one chunks in this setting, but do not
establish a universally optimal weight. We retain $\beta=5$
as the default and leave it configurable.

\subsection{Detailed Support-Selection Ablation}
\label{app:support-results}

Table~\ref{tab:support-ablation-complete} provides per-task
results for the support-selection ablation in
Section~\ref{sec:mismatch-support}. We compare three policies
for non-one-to-one chunks. Realized-only support retains only
the observed sequence pair for every such chunk. The proposed
policy uses full support for one-to-many and many-to-one chunks,
and realized-only support for many-to-many chunks. Full support
for all uses all retained admissible final-token pairs, with
the observed prefixes fixed, for every non-one-to-one chunk.
The one-to-one construction is unchanged, and all policies
include a residual event.

\begin{table*}[t]
  \caption{Detailed support-selection ablation
    with the Llama-3.2-1B student. The policies differ only in
    their treatment of non-one-to-one chunks. Avg.-5 is the unweighted mean over the five
    tasks. Time denotes separately measured average
    training-step time in seconds.
    \textbf{Bold} marks the highest Avg.-5 for each teacher.}
  \label{tab:support-ablation-complete}
  \centering
  \small
  \renewcommand{\arraystretch}{0.95}
  \resizebox{\linewidth}{!}{%
  \begin{tabular}{lrrrrrrr}
    \toprule
    \textbf{Support policy}
      & \textbf{MMLU} & \textbf{WG}
      & \textbf{GSM8K} & \textbf{MATH} & \textbf{MBPP}
      & \textbf{Avg.-5} & \textbf{Time (s)} \\
    \midrule
    \multicolumn{8}{l}{%
      \textbf{Qwen3-4B-Base
      $\rightarrow$ Llama-3.2-1B}} \\
    \addlinespace
    Realized-only for all
      & 45.16 & 61.80 & 21.61 & 7.94 & 31.40
      & 33.58 & 7.45 \\
    Proposed
      & 45.22 & 61.80 & 21.30 & 8.26 & 32.60
      & \textbf{33.84} & 7.45 \\
    Full support for all
      & 45.02 & 61.48 & 22.59 & 8.38 & 31.00
      & 33.69 & 21.31 \\
    \midrule
    \multicolumn{8}{l}{%
      \textbf{Phi-4-mini
      $\rightarrow$ Llama-3.2-1B}} \\
    \addlinespace
    Realized-only for all
      & 46.55 & 63.54 & 24.94 & 9.08 & 32.00
      & 35.22 & 6.99 \\
    Proposed
      & 46.47 & 62.98 & 26.91 & 9.22 & 33.60
      & \textbf{35.84} & 7.06 \\
    Full support for all
      & 46.09 & 64.17 & 25.40 & 9.08 & 33.00
      & 35.55 & 27.64 \\
    \bottomrule
  \end{tabular}%
  }
\end{table*}

\begin{table}[t]
  \caption{Teacher top-$k$ results with Qwen3-4B and Llama-3.2-1B.
  The full-logit setting uses no teacher-side truncation. Avg.-5 is the
  unweighted mean over the five tasks; time denotes average
  training-step time in seconds. \textbf{Bold} marks the highest
  average.}
  \label{tab:topk-complete}
  \centering
  \scriptsize
  \setlength{\tabcolsep}{3pt}
  \renewcommand{\arraystretch}{0.95}
  \resizebox{0.95\linewidth}{!}{%
  \begin{tabular}{l*{7}{N}}
    \toprule
    \textbf{Teacher support}
      & \textbf{MMLU} & \textbf{WG}
      & \textbf{GSM8K} & \textbf{MATH} & \textbf{MBPP}
      & \textbf{Avg.-5} & \textbf{Time (s)} \\
    \midrule
    $k=16$
      & 44.91 & 62.75 & 19.71 & 8.30 & 30.60 & 33.25 & 3.60 \\
    $k=128$
      & 45.16 & 62.60 & 21.15 & 8.24 & 30.80 & 33.59 & 3.66 \\
    $k=8192$
      & 44.77 & 62.83 & 21.61 & 8.48 & 30.60 & 33.66 & 4.03 \\
    \midrule
    Full logits
      & 45.22 & 61.80 & 21.30 & 8.26 & 32.60
      & \textbf{33.84} & 7.45 \\
    \bottomrule
  \end{tabular}%
  }
\end{table}

The proposed policy achieves the highest Avg.-5 and MBPP
score with both teachers. Compared with realized-only support,
it raises Avg.-5 from $33.58$ to $33.84$ with Qwen and from
$35.22$ to $35.84$ with Phi, at nearly unchanged step time.
Using full support for many-to-many chunks produces mixed
task-level changes. With Qwen, it increases GSM8K and MATH
scores relative to the proposed policy, but lowers MBPP
from $32.60$ to $31.00$, resulting in a lower Avg.-5.

Full support for all also requires $2.86\times$ and
$3.92\times$ as much time per step as the proposed policy
with Qwen and Phi, respectively. These results favor using
full support for one-to-many and many-to-one chunks while
retaining realized-only support for many-to-many chunks
in the evaluated settings.

\subsection{Detailed Teacher Top-$k$ Results}
\label{app:topk-results}

Table~\ref{tab:topk-complete} provides per-task scores and training-step
times for the configurations in Section~\ref{sec:topk-efficiency}.
Using Qwen3-4B and Llama-3.2-1B, we compare teacher-side truncation
at $k=16$, $128$, and $8192$ with full logits. All other settings
remain fixed.

All three top-$k$ settings remain within $0.59$ average points
of full-logit performance while reducing training-step time.
At $k=128$, Avg.-5 is $33.59$, only $0.25$ points below full
logits, and step time falls from $7.45$ to $3.66$ seconds,
a $2.04\times$ speedup. Increasing $k$ to $8192$ raises Avg.-5
to $33.66$ and step time to $4.03$ seconds. Individual task scores
are not monotonic in $k$, but aggregate accuracy remains close
to the full-logit result. These observations support teacher
top-$k$ truncation as an effective accuracy--efficiency trade-off
in this setting.
\vspace{-0.2in}
\paragraph{Timing protocol.}
The production timing benchmarks use $256$ NVIDIA H100 GPUs
across $32$ nodes, with a per-GPU microbatch size of $1$,
gradient accumulation of $3$, a global batch size of $768$,
and a sequence length of $4096$.
Teacher predictions are computed online.
Measurements include teacher and student forward passes,
loss and support computation, backward propagation,
distributed communication, optimizer updates, and data
loading and transfer. They exclude offline data preparation
and tokenizer-table construction, process and model startup,
checkpoint writing, and additional measurement synchronization.

\end{document}